\documentclass[a4paper,twoside]{article}

\usepackage{epsfig}
\usepackage{subcaption}
\usepackage{calc}
\usepackage{amssymb}
\usepackage{amstext}
\usepackage{amsmath}
\usepackage{amsthm}
\usepackage{multicol}
\usepackage{pslatex}
\usepackage{apalike}
\usepackage{SCITEPRESS}     

\begin{document}

\title{Multimodal Personality Recognition using Cross-Attention Transformer and Behaviour Encoding}

\author{\authorname{Tanay Agrawal\sup{1}, Dhruv Agarwal\sup{1, 3}, Michal Balazia\sup{1, 2}, Neelabh Sinha\sup{1, 4}, and François Bremond\sup{1, 2}}
 \affiliation{\sup{1}INRIA Sophia Antipolis - Méditerranée, France}
 \affiliation{\sup{2}Université Côte d'Azur, France}
 \affiliation{\sup{3}Indian Institute of Information Technology, Allahabad, India}
 \affiliation{\sup{4}Birla Institute of Technology and Science, Pilani, India}
 \email{\{firstname.secondname\}@inria.fr}
 }


\abstract{Personality computing and affective computing have gained recent interest in many research areas. The datasets for the task generally have multiple modalities like video, audio, language and bio-signals. In this paper, we propose a flexible model for the task which exploits all available data. The task involves complex relations and to avoid using a large model for video processing specifically, we propose the use of behaviour encoding which boosts performance with minimal change to the model. Cross-attention using transformers has become popular in recent times and is utilised for fusion of different modalities. Since long term relations may exist, breaking the input into chunks is not desirable, thus the proposed model processes the entire input together. Our experiments show the importance of each of the above contributions.}

\onecolumn \maketitle \normalsize \setcounter{footnote}{0} \vfill

\section{Introduction}
Personality is a combination of behavior, emotion, motivation, and thought patterns. Our personality greatly impacts our lives, defining choices, health along with our preferences and desires. Personality traits define a particular way of thinking, feeling, and behaving. Specifically, personality traits have been defined pertaining to individual well being and social-institutional outcomes like occupational choices, interpersonal relations, and success in various scenarios.

We make decisions using a two system model: rational and emotional. So, modelling the latter will help us build more accurate AI systems when it is coupled with the vast amount of research done on the former. The problem of personality recognition is complex and thus would require a lot of training data to get models usable in real-life. This is one
reason that multimodal learning is very popular in this domain. First Impressions v2 \cite{Junior_2021_WACV} is a multimodal dataset for personality recognition and is used in this work. We utilise all the information available in the dataset -- speech, body language, expressions and their surroundings along with their demographic information -- and define a new behaviour encoding to facilitate learning. Deep learning backbones have been found to extract meaningful features. Generally, larger models give better features. Due to the high number of inputs, it is not possible to have large backbones for all. So we decide to compute the additional behaviour encoding to have better features even with a smaller backbone. In multimodal learning, we need to process each modality individually and also find how they are correlated. Thus, we also show how to merge the behaviour encoding with an existing baseline \cite{palmero2021context}. Temporal processing is also important, we use LSTMs to have a higher temporal resolution. Even a simple temporal processing model helps as there are multiple modalities and the embedding input to the temporal processor is very rich in information.

For defining personality, the big five personality traits are used. They are often referred to as OCEAN: Openness, Conscientiousness, Extroversion, Agreeableness, and Neuroticism. These five traits represent broad domains of human behaviour and account for differences in both personality and decision making. Today, a major use of this model is by Human Resource practitioners to evaluate potential employees and marketers to understand the audiences of their products.

The focus on this problem is relatively new and there is limited work done till date. As discussed above, the task is complex so it requires careful processing of each input modality and their relations. Previous works either involves using a subset of modalities or only simple fusion techniques to fuse the modalities \cite{aslan2019multimodal}.

Another challenge in this domain is that annotations are generally provided by the participants through questionnaires or an online answering platform during or after the experiment sessions. They may also be annotated by third-party annotators, but personality is subjective so they are not always perfect. This makes the task even harder and further requires a method that can utilise all available modalities and formulate complex relations not only for each modality but also across modalities. Defining handcrafted inputs increases performance as there might be more direct correlation between them and other modalities or even the output as compared to the original inputs.

As stated above, ChaLearn First Impressions V2 challenge dataset \cite{snjsnsi} which is publicly available is used \cite{palmero2021context} for this work. The dataset consists of 10,000 videos of people facing and speaking to a camera. Videos are extracted from YouTube, they are mostly in high-definition ($1280\!\times\!720$ pixels), and, in general, they have an average duration of 15 seconds with 30 frames per second. In the videos, people talk to the camera in a self-presentation context and there is a diversity in terms of age, ethnicity, gender, and nationality. The videos are labeled with personality factors using Amazon Mechanical Turk (AMT), so the ground truth values are obtained by using human judgment. For the challenge, videos are split into training, validation and test sets with a 3:1:1 ratio and we choose to use the same to compare the results.

Summarising our contributions in this work, we introduce a handcrafted behaviour embedding that improves performance and reduces convergence time of the model. We modify the chosen baseline to incorporate new modalities (transcript and behaviour encoding) and also address missing temporal relations in it. We also achieve state of the art results for personality recognition on the chosen dataset.





\section{Related Work}
This section discusses the work done on personality recognition using different techniques and modalities. They can be broadly classified into the following categories.

\subsection{Using Video}
As in the case of most visual deep learning tasks, Convolutional Neural Networks~(CNNs) are the most commonly used in the field of personality detection. Facial attributes can be an important factor in predicting social traits  \cite{qin2016modern,VernonE3353}. Impressions that influence people’s behavior towards other individuals can be accurately predicted from videos \cite{10.1007/978-3-319-49409-8_30}. Many researchers have experimented with different ways of capturing facial features such as in the form of Facial Action Coding System~(FACS) which extracts action units such as raised eyebrows or blinking, and morphological features \cite{8265578}.

\subsection{Using Audio}
Using audio as the only input modality is not a popular choice for personality recognition. It is combined with video in most of the cases resulting in bimodal approaches.  In the existing ones, audio features like Mel-Frequency Cepstral Coefficients~(MFCC), Zero Crossing Rate~(ZCR), Logfbank, other cepstral and spectral ones serve as inputs into regressors. Analyzing conversations \cite{valente12_interspeech} and the pitch, timber, tune and rhythm of the voice \cite{123456}, it is possible to recognize the personality traits or predict the speaker attitudes automatically. These approaches demonstrate that audio information is important for personality.

\subsection{Using Text}
Looking at the textual modality, preprocessing is an important step. Generally, extracted features include Linguistic Inquiry and Word Count~(LIWC) \cite{mikolov2013distributed}, Mairesse, Medical Research Council~(MRC), which are then fed into standard classifiers or regressors. Learning word embeddings and representing them as vectors, either with GloVe, Word2Vec, GRU, LSTM or recently BERT, is also a very commonly followed approach. It was observed that combining text features with something else such as metadata and convolutions results in better performance paving the path to multimodal approaches. 

Social networks provide rich textual data for the recognition of personality traits \cite{bosbdd,article123}. Transcribed videos blogs and dialogues also provide useful information for this task \cite{10.1145/2659522.2659530}.

\subsection{Multimodal approaches}
Personality traits can be detected in self presentation videos based on the acoustic and visual, non-verbal features such as pitch, intensity, movement, head orientation, posture, fidgeting and eye-gaze. Zheng et al. \cite{4468714} shows body gestures, head movements, expressions, and speech lead to effective assessment of personality and emotion. According to Sarkar et al. \cite{10.1145/2659522.2659528}, features such as audiovisual, text, demographic and sentiment features are important for our task.

Although multimodal approaches are commonly used to recognize personality traits, there does not exist a comprehensive method utilizing a considerable amount of informative features. Most of the multimodal approaches perform late fusion. Deep bimodal regression give state of the art results  \cite{he2015deep}. Some other approaches with good results are \cite{7899605, inbook} and \cite{8066355}. Each modality features may be used together for personality prediction, this approach is called early fusion. Present research in the field aims to find efficient ways of feature extraction and combination. Few models which have dealt with trimodal fusion of features \cite{aslan2019multimodal,palmero2021context}. Emotion recognition is a closely related problem and has interesting approaches for multimodal data processing \cite{dai2021multimodal,tsai2019MULT}. Our approach aims to utilise all possible information available and also some extra features computed similar to the ideas discussed in the beginning of this subsection.

\section{The Proposed Framework}

\begin{figure*}
\includegraphics[width=\textwidth]{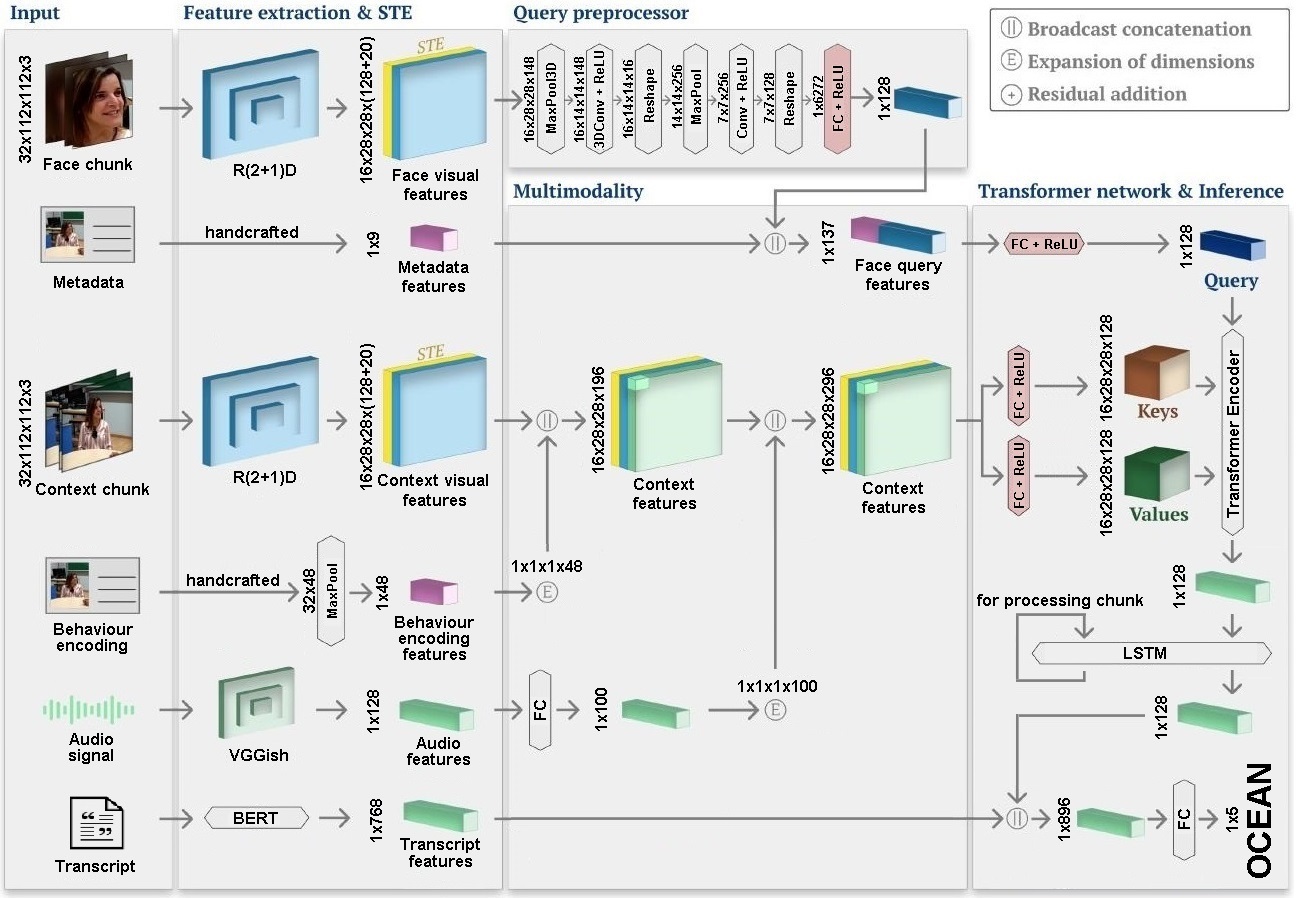}
\caption{Proposed method to infer self-reported personality (OCEAN) traits from multimodal data. Input consists of visual (face and context chunks), audio (raw chunks), metadata of the target person, handcrafted behaviour encoding and transcript of the audio. Feature extraction is performed by a R(2+1)D network for the visual chunks, VGGish for audio and BERT for the transcript. The visual features from the R(2+1)D’s 3rd residual block are concatenated to spatiotemporal encodings (STE). The VGGish’s audio features and handcrafted metadata features are incorporated to visual context/query features and the result transformed to the set of Query, Keys, and Values as input to the Transformer encoder. The output of the transformers are sequentially passed to an LSTM chunkwise. The transcript features from BERT are concatenated with these and finally  fed to a fully-connected (FC) layer to regress per-video OCEAN scores.}
\label{fig1}
\end{figure*}

The approach uses face crops of the target person and relates it to body language, surroundings and speech using a transformer based architecture. Short-term temporal relations are processed in this way and longer temporal relations are established using LSTM. For transcript analysis, short term temporal relations are not very meaningful so the features for the entire input sequences are extracted using BERT \cite{devlin2019bert}. Late fusion is then finally used for inferring the OCEAN personality traits. There are several stages in the proposed method and they are discussed in the following sections. Figure~\ref{fig1} shows the overview of the entire architecture.

\subsection{Preparing the Input and Feature Extraction}
The audiovisual data is pre-processed in a similar manner as in \cite{palmero2021context}. 32 frames with a stride of 2 are taken for video based inputs and R(2+1)D \cite{8578773} is used to extract spatio-temporal features. Stride is modified depending on the frame rate of the video to keep the time span of the chunks roughly the same. Audio clip with the same time span is converted to a tensor for input in the same way as in the method used in VGGish \cite{45611}. BERT is used for extracting features of the transcript. The method for computing behaviour encoding is discussed in the next section in detail. Demographic data -- age, gender, ethnicity and attractiveness -- are also used. The value are either one hot encode or normalised to the range $[0,1]$. Table~\ref{tab:ablation_results} and \cite{escalante2019explaining} give the details of each element of the feature vector.

Note that Attractiveness is only available for people with Caucasian ethnicity, but since $\sim\!\!\!86\%$ of the people in the dataset are Caucasian, it is utilised and the default Attractiveness value of 0 is set for people of other ethnicities.    

\begin{table}[h!]
\centering
\begin{tabular}{|c|c|}
\hline
Demographic variable & Dimension \\
\hline
Ethnicity & 3D (one hot encoding)\\ 
Gender & 2D (one hot encoding)\\ 
Age & 1D\\
Attractiveness & 1D\\
\hline
\end{tabular}
\caption{Dimensions of demographic metadata}
\label{tab:ablation_results}
\end{table}
\subsection{Behaviour Encoding}
We compute behaviour encoding for 13 actions: head tilt, thrust, bob, lips in, mouth corner, frown, small mouth, wrinkle, crouch, lean forward, fold arms, hand to face, hand to mouth. For detecting the individual behaviors, we use a rule based approach on the skeleton and facial key points. For extracting the key points (skeleton and face), we use LCRNet \cite{Rogez_2019} and OpenFace \cite{openface}.

\begin{table}[ht]
\small
\centering
\tabcolsep1.5pt
\begin{tabular}{|l|p{11.0em}|l|l|}
\hline			
behavior & extracted feature $x$ & $c_\sigma$ & $\lambda_\sigma$ \\\hline
head tilt & head roll angle & $10$ & $1$ \\\hline
thrust & derivative of translation vector along $z$ axis when derivative along other directions is less than $10$ and direction of derivative in previous and next frame is the same & $-25cm/s$ & $1$ \\\hline
bob & derivative of pitch angle when derivative of yaw angle is less than $20$ and direction of derivative in previous and next frame is the same
 & $-50deg/s$ & $1$ \\\hline
lips in & FACS action unit Lip Suck & - & - \\\hline
mouth corner & FACS action unit Lip Stretcher & $1.2$ & $6$ \\\hline
frown & FACS action unit Brow Lowerer & $1.2$ & $6$ \\\hline
small mouth & FACS action unit Lip Tightener & $1.2$ & $6$ \\\hline
wrinkle & FACS action unit Nose Wrinkler & $1.2$ & $6$ \\\hline
crouch & distance between knees and head & $30cm$ & $-0.35$ \\\hline
lean forward & $z$ coordinate on distance between root and shoulders & $10cm$ & $4$ \\\hline
fold arms & alternate distance between elbows and wrists when $y$ coordinate of both elbows are less than $10cm$ & $20cm$ & $-0.5$ \\\hline
hand to face & distance between wrists and head & $35cm$ & $-0.5$ \\\hline
hand to mouth & distance between wrists and head minus $10cm$ on $y$ axis & $25cm$ & $-0.5$ \\\hline
\end{tabular}
\newline
\caption{Detection methods of 13 behaviors.}
\label{t-beh}
\end{table}

In each frame, we infer a detection confidence for all behaviors in the scale of 0--1, where 1 represents complete confidence in presence of the behavior and 0 represents complete confidence in absence of the behavior. This is done by extracting a specific feature $x$ and transforming it through a sigmoid function
\[f_\sigma\left(x\right)=\frac{1}{1+e^{-\lambda_\sigma\left(x-c_\sigma\right)}}\]
with parameters of center $c_\sigma$ and multiplier $\lambda_\sigma$. As shown in Table~\ref{t-beh}, each behavior is characterized by a specific extracted feature and the two sigmoid function parameters.

\subsection{Positional Encoding for the Transformer}
Positional encodings are important to be added to the input with transformer based models as they make the model order invariant. Sinusoidal encodings are common, but we choose to use learned encodings in our experiments. As we need to process in both space and time, we need an encoding for both. We initialize encodings for both and use a two layer fully connected network for learning them. Then they are broadcast concatenated to each other resulting in an encoding which can be concatenated to the input.
\subsection{Preparing Inputs for the Transformer}
Features extracted from face crops of the complete frame input are further processed and are used as the query for the transformer. To factor its relation with the rest of the information in the complete frame and audio inputs, they are processed to be used as key and value. The face features are passed through the following layers to get the input query:

\begin{enumerate}
\item 3D max pooling layer with a kernel size and stride of 2 for height and width dimensions, and 1 for the temporal dimension.
\item 3D convolution layer with kernel size of 1 for all dimensions and 16 kernels.
\item ReLU activation followed by reshaping to merge temporal and channel dimensions.
\item 2D max pooling layer with a kernel size and stride of 2 for height and width dimension.
\item 2D convolution layer with kernel size of 2 for all dimensions and 128 kernels.
\item ReLU activation followed by flattening.
\item A fully connected layer to change the shape to 128, followed by ReLU activation and dropout $p=0.2$ layer.
\end{enumerate}

Demographic metadata is concatenated to the obtained feature vector and is passed through a fully connected and ReLU layer to obtain a 128 dimensional query vector.

Behaviour encoding is broadcast concatenated to complete frame features which already contain spatio-temporal positional encoding. The audio features are projected into a 100 sized feature vector using a fully connected layer and broadcast concatenated with the above obtained complete frame features. These are passed through separate fully connected and ReLU layers to obtain keys and values for the transformer.
\subsection{Transformer, Temporal Processing and Fusion with Transcript Features}
The transformer consists of only the encoder with 2 attention heads and stacked 3 times, that is, 3 layers. The hidden dimension is 128. The transformer processes roughly $2.5$ seconds of the input in one forward pass. These chunks are passed through two stacked LSTM blocks to find long-term temporal relations. The hidden state after the last chunk is passed, is concatenated with transcript features and passed through linear, ReLU and dropout layers to obtain the 5 personality trait values.
\section{Experiments and Results}
\subsection{Training Details}
To reduce training time, the parameters of backbones R(2+1)D, VGGish and BERT are frozen and are not updated during backpropagation. There is one exception to this, we finetune our model with the weights unfrozen as behaviour encoding helps improve the performance of the backbone also and to exploit that, we finetune our model for 20 epochs. One RTX 6000 GPU is used for training and the batch size is set to 8. Learning rate used is $10^{-5}$ with the scheduler "ReduceLRonPlateau", patience 5 and factor 0.5. Further details of the experiments are given in section 4.3. Figure 2 and 3 show training graphs of two different experiments, an ablation study with the proposed framework without transcript and the proposed framework, respectively. It is interesting to note that adding modalities decreases the number of epochs required for convergence as shown by these two figures.

\begin{figure}
\begin{center}
  \epsfig{file=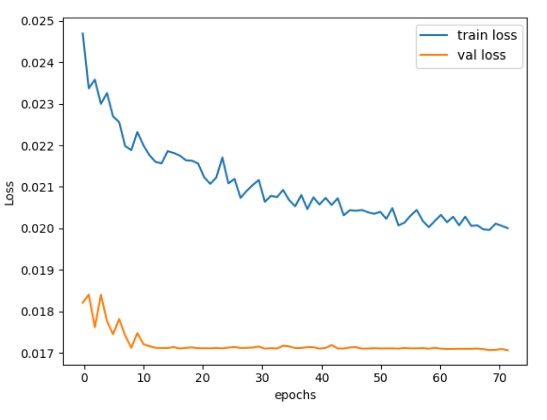, width=\columnwidth}
  \caption{MSE Loss curves for w/o Transcript Ablation experiment}
\label{fig:Loss_noTBehav}
\end{center}
\end{figure}

\begin{figure}
\begin{center}
  \epsfig{file=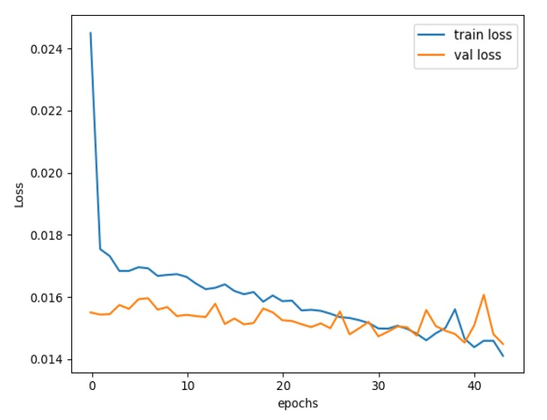, width=\columnwidth}
  \caption{MSE Loss curves for our proposed approach}
\label{fig:Loss_noTBehav}
\end{center}
\end{figure}

\subsection{Evaluation Protocol}
The evaluation metric is chosen to be the same as that of the ChaLearn challenge where the dataset was released. The OCEAN traits have five classes which are rated in the range $[0,1]$. The challenge \cite{snjsnsi} defines mean accuracy $A$ over all predicted personality trait values as
\begin{equation}
\label{eq:eg}
A = 1 - {\frac{1}{N}\sum_{i=1}^{N}\left|t_i - p_i\right|}
\end{equation}
where $t_i$ are the ground truth scores and $p_i$ are the predicted scores for personality traits summed over $N$ videos.

\subsection{Results and Ablation Studies}
We compare our result to the previous state of the art and also perform ablation studies to show the need for all modalities present. Table 3 enumerates these results. 

\begin{table}[ht]
\small
\centering
\tabcolsep1.5pt
\begin{tabular}{|p{7.75em}|l|l|l|l|l|l|}
\hline
\textbf{Model} & \multicolumn{6}{l|}{\textbf{Accuracy}}\\\hline\hline
 & O & C & E & A & N & Mean\\\hline
Aslan and  \cite{aslan2019multimodal} & .9166 & .9214 & .9208 & .9189 & .9162 & .9188 \\\hline
DCC \cite{G_l_t_rk_2016} & .9117 & .9113 & .9110 & .9158 & .9091 & .9122 \\\hline
evolgen \cite{Subramaniam2016BimodalFI} & .9130 & .9136 & .9145 & .9157 & .9098 & .9138 \\\hline
Gurpinar et al. \cite{7899605} & .9141 & .9141 &.9186 & .9143 &.9123 & .9147 \\\hline
PML \cite{8014945} & .9138 & .9166 & .9175 & .9166 & .9130 & .9170 \\\hline
BU-NKU \cite{8014944} & .9169 & .9166 & .9206 & .9161 & .9149 & .9170 \\\hline
Our proposed model & \textbf{.9291} & \textbf{.9258} & \textbf{.9272} & \textbf{.9288} & .9210 & \textbf{.9263}\\\hline\hline
Baseline: w/o behaviour encoding and transcript & .8959 & .8996 & .8987 & .8938 & .8932 & .8962
\\\hline
w/o behaviour encoding & .9095 & .9094 & .9112 & .9133 & .9041 & .9095\\\hline
w/o transcript & .9013 & .8992 & .8988 & .9041 & .8996 & .9006\\\hline
w/o LSTM & .8892 & .8532 & .9131 & .9024 & \textbf{.9315} & .8978\\\hline
w/o metadata & .9260 & .9212 & .9234 & .9249 & .9168 & .9225 \\\hline
\end{tabular}
\newline
\caption{Experiments and Results; O: Openness, C: Conscientiousness, E: Extroversion, A: Agreeableness, and N: Neuroticism}
\label{t-beh}
\end{table}

We achieve state of the art results as we utilise all the available information and also compute an additional behaviour embedding to facilitate learning. This method of computing a behaviour encoding can be utilised in a variety of use-cases and we predict that it will help in reducing training time and improving results in other areas, such as action recognition also.

The ablation study proves the efficacy of our approach, showing the importance of using different input modalities and the difference in results is significant. All the different models discussed below are trained in parallel and not sequentially, that is, the later models were not finetuned from the initial ones. The first approach includes the baseline model and has the same inputs modified as per the dataset details. The baseline has all the inputs except the behaviour encoding and the transcript. This experiment is to establish our own baseline results to compare against.

We see the results of the model without behaviour encoding. There is roughly $1.8\%$ decrease in accuracy which shows that behaviour encoding facilitates in the prediction of personality.

We also observe the performance of the model without transcript. This shows a similar trend as behaviour encoding - there is a slightly less decrease in accuracy but the difference is minute.

For finding the performance of the model with LSTM, we keep everything the same but take the median value across chunks for each video to get the output. Without LSTM, the model behaves erratically. As expected the accuracy decreases for most classes. But, for the class neuroticism, the best results are without LSTM. One explanation is the high variability in the inputs where neuroticism is high and the LSTM which tries to identify a pattern across chunks does not perform very well.

The last experiment is without metadata about the target person. There is not much difference in results as compared to the other inputs, but we still see a reduction in performance. So, demographic data about a person affects personality too. Some bias in the data is the most probable cause but since the dataset is large, in our opinion this is not the case and the inference drawn holds.
\section{Conclusions}
In this work, we show that a model for personality recognition will benefit from more modalities and data as input. We propose a new handcrafted behaviour encoding where each element is the probability of a low level action relevant to the task. We show the effectiveness of all the inputs in the data through ablation studies. We also give our opinion on the trends shown in the ablation studies. Owing to the interdisciplinary nature of the project, there are numerous additions that will further improve performance. From intuition, there are some which might improve performance by a higher margin than others. Using better backbones for feature extraction would be interesting. We use the same ones as in the baseline we choose but there are existing models with better performance for similar tasks that can be utilised. Transformers have been shown to perform better than LSTMs. In the future, we will try to increase temporal scale of attention in the transformer rather than using a separate module for combining information across chunks. This might tackle the problem that is seen with neuroticism as discussed in section 4.3. One of the major drawbacks of multimodal data is that preprocessing takes a lot of time. Thus, it will be interesting to explore Knowledge Distillation to allow the model to utilise one or a subset of modalities and give a similar performance but with lesser inputs. We would also like to test our approach on other big scale multimodal datasets, when they are available in the future. This area of work has a lot of applications in healthcare which we are exploring and hope that this work leads to advancement in the area. We also hope that it motivates other people to work on this interesting problem.

\bibliographystyle{apalike}
{\small
\bibliography{paper}}

\end{document}